\documentclass[sigconf,authorversion]{acmart}

\AtBeginDocument{%
  \providecommand\BibTeX{{%
    \normalfont B\kern-0.5em{\scshape i\kern-0.25em b}\kern-0.8em\TeX}}}

\setcopyright{acmcopyright}
\copyrightyear{2018}
\acmYear{2018}
\acmDOI{XXXXXXX.XXXXXXX}

\acmConference[Conference acronym 'XX]{Make sure to enter the correct
  conference title from your rights confirmation emai}{June 03--05,
  2018}{Woodstock, NY}
\acmPrice{15.00}
\acmISBN{978-1-4503-XXXX-X/18/06}

\usepackage{subcaption}
\begin{document}

\title{ Regions are Who Walk Them: a Large Pre-trained Spatiotemporal Model Based on Human Mobility for Ubiquitous Urban Sensing}


\author{Ruixing Zhang}
\affiliation{%
  \institution{SKLSDE Lab, Beihang University}
  \city{Beijing}
  \country{China}}
\email{yyxzhj@buaa.edu.cn}

\author{Liangzhe Han}
\affiliation{%
  \institution{SKLSDE Lab, Beihang University}
  \city{Beijing}
  \country{China}}
\email{hanlz@buaa.edu.cn}

\author{Leilei Sun}
\affiliation{%
  \institution{SKLSDE Lab, Beihang University}
  \city{Beijing}
  \country{China}}
\email{leileisun@buaa.edu.cn}

\author{Yunqi Liu}
\affiliation{%
  \institution{SKLSDE Lab, Beihang University}
  \city{Beijing}
  \country{China}}
\email{liuyunqi@buaa.edu.cn}

\author{Jibin Wang}
\affiliation{%
  \institution{China Mobile Information Technology Center}
  \city{Beijing}
  \country{China}}
\email{wangjibin@chinamobile.com}

\author{Weifeng Lv}
\affiliation{%
  \institution{SKLSDE Lab, Beihang University}
  \city{Beijing}
  \country{China}}
\email{lwf@buaa.edu.cn}

\begin{abstract}
User profiling and region analysis are two tasks of significant commercial value.
However, in practical applications, modeling different features typically involves four main steps: data preparation, data processing, model establishment, evaluation, and optimization. This process is time-consuming and labor-intensive. Repeating this workflow for each feature results in abundant development time for individual tasks and a reduced overall volume of task development.
Indeed, human mobility data contains a wealth of information. Several successful cases suggest that conducting in-depth analysis of population movement data could potentially yield meaningful profiles about users and areas. Nonetheless, most related works have not thoroughly utilized the semantic information within human mobility data and trained on a fixed number of the regions.
To tap into the rich information within population movement, based on the perspective that \textbf{R}egions \textbf{A}re \textbf{W}ho walk them, we propose a large spatiotemporal model based on trajectories (RAW). It possesses the following characteristics: 1) Tailored for trajectory data, introducing a GPT-like structure with a parameter count of up to 1B; 2) Introducing a spatiotemporal fine-tuning module, interpreting trajectories as collection of users to derive arbitrary region embedding. This framework allows rapid task development based on the large spatiotemporal model.
We conducted extensive experiments to validate the effectiveness of our proposed large spatiotemporal model. It's evident that our proposed method, relying solely on human mobility data without additional features, exhibits a certain level of relevance in user profiling and region analysis. Moreover, our model showcases promising predictive capabilities in trajectory generation tasks based on the current state, offering the potential for further innovative work utilizing this large spatiotemporal model. The source code will be released at \url{www.github.com/Rising0321/RAW}.
\end{abstract}


\maketitle

\section{Introduction}

Profiling users and analyzing regions are two tasks of significant business value. In today's digital age, businesses and organizations increasingly rely on data to understand their customers and markets better. By profiling users, businesses can gain a deeper understanding of their customers' needs, preferences, and behaviors, enabling them to tailor products and services more accurately, thereby increasing customer satisfaction, sales, and loyalty \cite{user_profiling_1,user_profiling_2,user_profiling_3}. Simultaneously, analyzing regions can reveal market trends, competitive landscapes, and opportunities in different areas, helping businesses plan market strategies, allocate resources, and expansion plans more effectively \cite{region_embed_1}. The two tasks altogether provides profound insights, contributing to sustainable business growth and competitive advantage.

In the industry, trying to model a feature in profiling users or analyzing regions typically involves four time-consuming steps. First, the data preparation phase necessitates the collection, cleaning, and organization of extensive user data to ensure its quality and consistency. This often demands significant time and resources. Subsequently, the data processing stage involves complex operations such as feature extraction, dimensionality reduction, and standardization to make the data suitable for modeling. However, this requires strong domain knowledge to identify truly valuable features. The model development phase entails algorithm selection and training, demanding robust algorithmic thinking and programming skills. Finally, model evaluation and optimization constitute an iterative process that involves continuous parameter adjustments and performance validation to ensure the model accurately reflects user characteristics and behaviors. Furthermore, each feature necessitates going through this entire process, resulting in prolonged individual task development times and an overall reduced capacity for task development.

Actually, human mobility data holds a wealth of information\cite{human_mobility_data}. For instance, if an individual regularly travels to and from specific locations on workdays, it is highly likely that he or she is a commuter. Similarly, if a person's residence and activities predominantly occur within a university campus, he or she is likely a student. If an area consistently experiences a substantial daytime outflow of people and an influx at night, it is likely to be a residence. Likewise, if a significant number of individuals frequently depart from and return to the same location while visiting tourist destinations in between, that place may be a hotel. These examples illustrate that by analyzing population mobility data, meaningful profiles of both users and regions can potentially be obtained.

\begin{figure}[H]
  \centering
  \includegraphics[width=\linewidth]{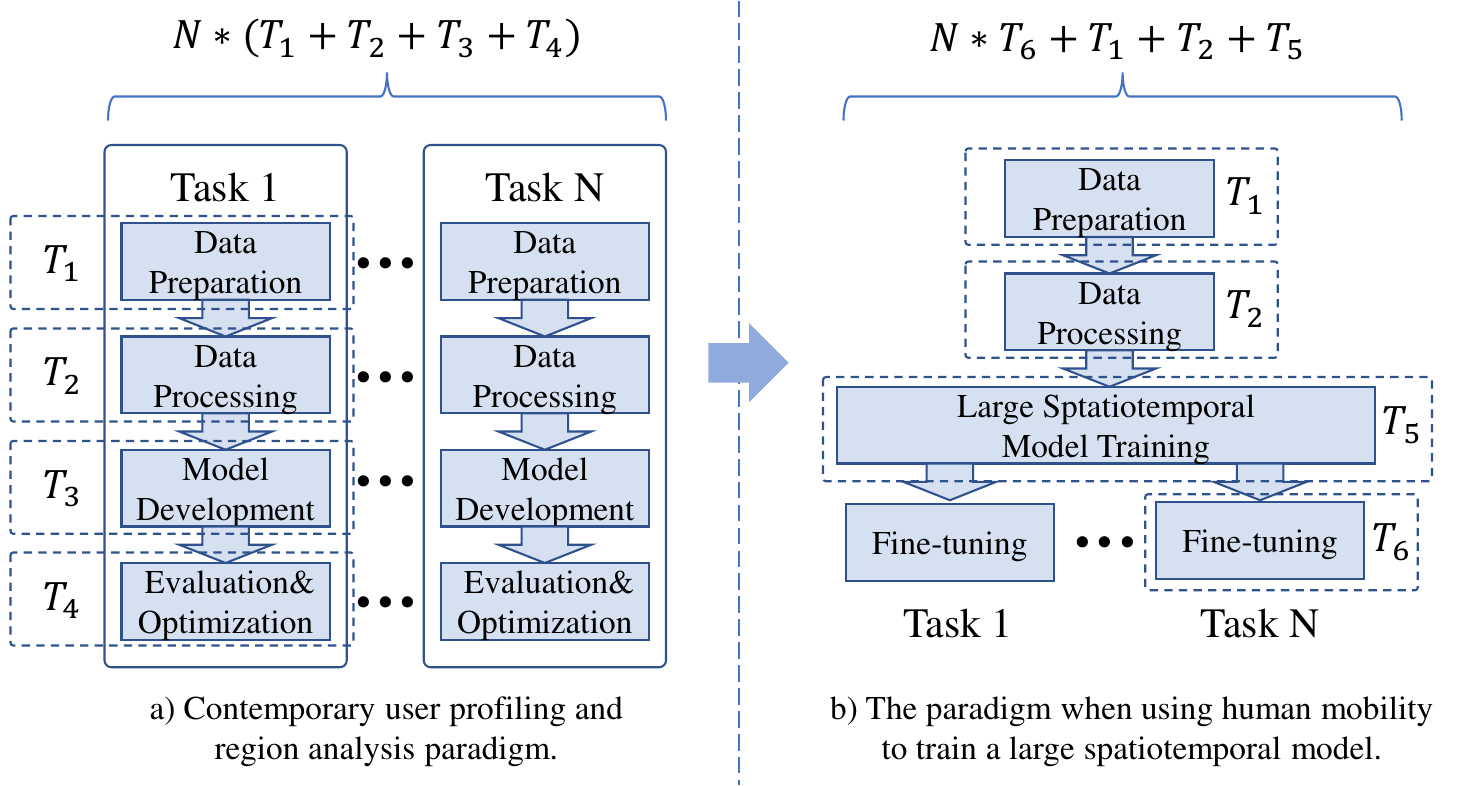}
  \caption{The change of user profiling and region analysis paradigm when using human mobility to train a large spatiotemporal model.}
\end{figure}

In recent years, human mobility data, as an important form of spatiotemporal data, has been widely utilized for profiling user and analysing region \cite{region_rep_1,region_rep_2,region_rep_3,region_rep_4}. These works have showcased the vast potential of human mobility data in mining spatiotemporal information. For instance, \cite{region_rep_1} utilized large-scale taxi mobility data to learn the characteristics and similarities among regions; \cite{region_rep_2} constructed a hypergraph network to represent relationships between regions by extracting features from mobility patterns; \cite{region_rep_3} introduced Points of Interest (POI) information and co-learning modules to enhance the semantic information of regions; \cite{region_rep_4} predicted and recommended regions based on metrics like in-degree and out-degree. However, these works also exhibit certain limitations: 1. They do not adequately consider the deeper semantic aspects of travel, merely retaining statistical features of travel; 2. They do not offer personalized embeddings for different users; 3. Their training is specifically tailored to particular tasks, making it challenging to swiftly adapt to various other tasks. 4. They can not generate embeddings for arbitrary given regions. 

Inspired by contemporary advancements in large model, particularly their efficient parallelization for sentence encoding, this paper endeavors to treat human behavioral trajectories as sentences to obtain pre-trained embeddings of these trajectories. This approach is motivated by several factors: 1. Sentences can be viewed as sequences of words, much like trajectories can be regarded as sequences of GPS coordinates. 2. Sentences can undergo pre-training in an autoregressive manner, and similarly, trajectories can be pre-trained using autoregressive methods. 3. Different segments of text can be trained in parallel, and likewise, different segments of trajectories can undergo parallel training. The subsequent natural question pertains to how to obtain embeddings for regions. In this paper, we posit that region are who walked them, which can be defined by the individuals traversing those areas. Therefore, we propose employing pooled embeddings of the trajectories of individuals traversing a particular region to derive embeddings for arbitrary region.

Therefore, we propose a large spatio temporal model based on a generative pre-trained transformer (GPT) for profiling users and analyzing regions. Our model comprises two components: pre-training and fine-tuning. In the pre-training phase, we initially encode the GPS coordinates of trajectories using a multi-layer perceptron (MLP), treating them as embeded tokens input into an L-layer transformer decoder layer. Subsequently, we utilize a linear layer to autoregressively predict the next GPS coordinate point. In the fine-tuning phase, we employ an MLP to predict the feature for users or regions. For user features, we input the user's trajectories into the pre-trained GPT to obtain their embedding as input. For region labels, we pool the embeddings of users who walked through an area over a specified period and input them to the MLP.

The contributions of this paper are as follow:

\begin{itemize}

\item We propose a large spatiotemporal model based on GPT. Through the autoregressive training process, each trajectory can be embeded into an informative embedding to fully represent the reflected human mobility. 

\item We introduce a spatiotemporal fine-tune mechanism. By simply applying a MLP on the trajectory embedding or the region embedding obtained by pooling, the upstream embedding can adaptive to the downstream tasks. Through our design, we can generate embeddings for arbitrary regions.

\item Extensive experiment including user profiling, region analysis proves the effectiveness of our proposed large spatiotemporal model. Moreover, the analysis of trajectory generation reflects the model learn the pattern of a human so that our model could have a broaded impact.

\end{itemize}

\section{Related Work}

\subsection{Deep Learning for Human Mobility}

In recent years, human mobility data, as an important form of spatiotemporal data, has been widely utilized for profiling user and analysing region \cite{region_rep_1,region_rep_2,region_rep_3,region_rep_4}. These works have showcased the vast potential of human mobility data in mining spatiotemporal information. For instance, \cite{region_rep_1} utilized large-scale taxi mobility data to learn the characteristics and similarities among regions; \cite{region_rep_2} constructed a hypergraph network to represent relationships between regions by extracting features from mobility patterns; \cite{region_rep_3} introduced Points of Interest (POI) information and co-learning modules to enhance the semantic information of regions; \cite{region_rep_4} predicted and recommended regions based on metrics like in-degree and out-degree. However, these works also exhibit certain limitations: 1. They do not adequately consider the deeper semantic aspects of travel, merely retaining statistical features of travel; 2. They do not offer personalized embeddings for different users; 3. Their training is specifically tailored to particular tasks, making it challenging to swiftly adapt to various other tasks. 4. They can not generate embeddings for arbitrary given regions.

\subsection{Large Model}

GPT-1\cite{GPT-1} was the first work investigated the potential of large language models in representation learning by modifying the framework of Transformer\cite{transformer}. It employed an autoregressive approach, learning sentence embeddings by predicting the next. The trained model could be fine-tuned for downstream tasks such as sentiment classification. GPT-2\cite{GPT-2} and GPT-3\cite{GPT-3} retained a similar model architecture of GPT-1 but significantly increased the model parameters and data scale, thereby achieving better results, proving the effectiveness of the GPT structure. 
However, despite the remarkable achievements of large language models in the field of natural language processing, the challenge remains in how to extend their applications to other domains. Some researchers\cite{protein} believe that the key to realizing large models lies in having readily available labels. Thus they have proposed methods of self-supervised learning to use a large number of labels in input itself. On the other hand, while large models have already achieved some success in language\cite{GPT-2,GPT-3}, image\cite{CLIP, dalle2}, and multimodal\cite{multimodal1,multimodal2} domains, achieving the same level of effectiveness in the spatiotemporal domain remains a difficult problem.

\section{Preliminaries}
\subsection{Trajectory}
We define a trajectory as a sequence composed of GPS coordinate points, denoted as \(T_i = \{C_{i,1},C_{i,2},\cdots,C_{i,|T_i|}\}\), where $i$ is the number of the trajectory, and each coordinate point \(C_{i,j} = \{x_{i,j},y_{i,j},t_{i,j}\}\) contains longitude, latitude, and time information, satisfying \(t_{i,j}<t_{i,j+1}\). It's important to note that the GPS coordinate points used in this paper do not represent the user's real location but are derived from the positions of the base stations where the user's phone communicates with, thus may have a certain deviation from the user's actual location. Additionally, in order to have uniformly spaced time sequences as inputs, we interpolate the data such that for any \(j>0\), \(t_{i,j}-t_{i,j-1}=15 mins\).
\subsection{Problem Definition}
Our main tasks can be categorized into two types: user profiling and region analysis. Both of the two tasks are input only with the trajectories. To provide a clearer definition of our tasks, we express user profiling andregion analysis in mathematical terms as follows:

\textbf{User Profiling}: Given a user's trajectory, denoted as $T_i$, and a task $q_j$, our objective is to predict the probability $p_{i,j}$ or value $r_{i,j}$ on that task. For instance, if $q_{i,j}$ represents whether the user is a commuter, $p_i$ would be the probability that the user is a commuter; if $q_j$ represents the number of taking subway, $r_{i,j}$ would be the number of taking subway.

\textbf{Region Analysis}: Given a set of users' trajectories, denoted as ${T_i}$ and a region-related task $q_j$, our goal is to predict the probability $p_{i,j}$ or value $r_{i,j}$ on that task. For example, if $q_j$ indicates whether the area is a shopping district, $p_{i,j}$ would represent the probability that the area is a shopping district; if $q_j$ represents the count of a specific type of Point of Interest (POI) nearby, $r_{i,j}$ would represent the count of that type of POI within the area.

\subsection{Next GPS prediction}
Since we use the next GPS prediction as the pretrain task, its definition is introduced here.

Given a trajectory $T_i = \{C_{i,1},C_{i,2},\cdots,C_{i,|T_i|}\}$. The goal of the Next GPS prediction task is to predict the next GPS coordinate $(x_{i,|T_i|+1},y_{i,|T_i|+1})$ based on the known trajectory. This task can be formally represented as finding the GPS coordinate $(x_{i,|T_i|+1},y_{i,|T_i|+1})$ that minimize the probability, given the trajectory $T_i$:
\begin{equation*}
\theta_{\text{pre}}^* = \arg\min_{\theta_{\text{pre}}} L_1\left(\left(x_{i,|T_i|+1},y_{i,|T_i|+1})\right), f_{\text{pre}}(T_i; \theta_{\text{pre}})\right)
\end{equation*}
where $\theta_{\text{pre}}$ denotes the parameters of the pre-train model.

\subsection{Fine-Tune}
Fine-tuning is a commonly used transfer learning method that allows for quick and effective learning on new tasks using pre-trained models. In this paper, we consider that the embeddings obtained from training on upstream tasks already contain rich and general information. Therefore, we employ a simple yet effective fine-tuning approach, where the embeddings obtained from the upstream task are used as input and fed into a Multi-Layer Perceptron (MLP) for prediction on downstream tasks, with only the MLP's parameters being updated. This fine-tuning approach can be formally defined in mathematical terms as follows:
\begin{equation*}
\theta_{\text{fine}}^* = \arg\min_{\theta_{\text{fine}}} L_2(y, f_{\text{fine}}(x; \theta_{\text{pre}}, \theta_{\text{fine}}))
\end{equation*}
where $\theta_{\text{fine}}$ denotes the parameters of the MLP.

\begin{figure*}[h]
  \centering
  \includegraphics[width=\linewidth]{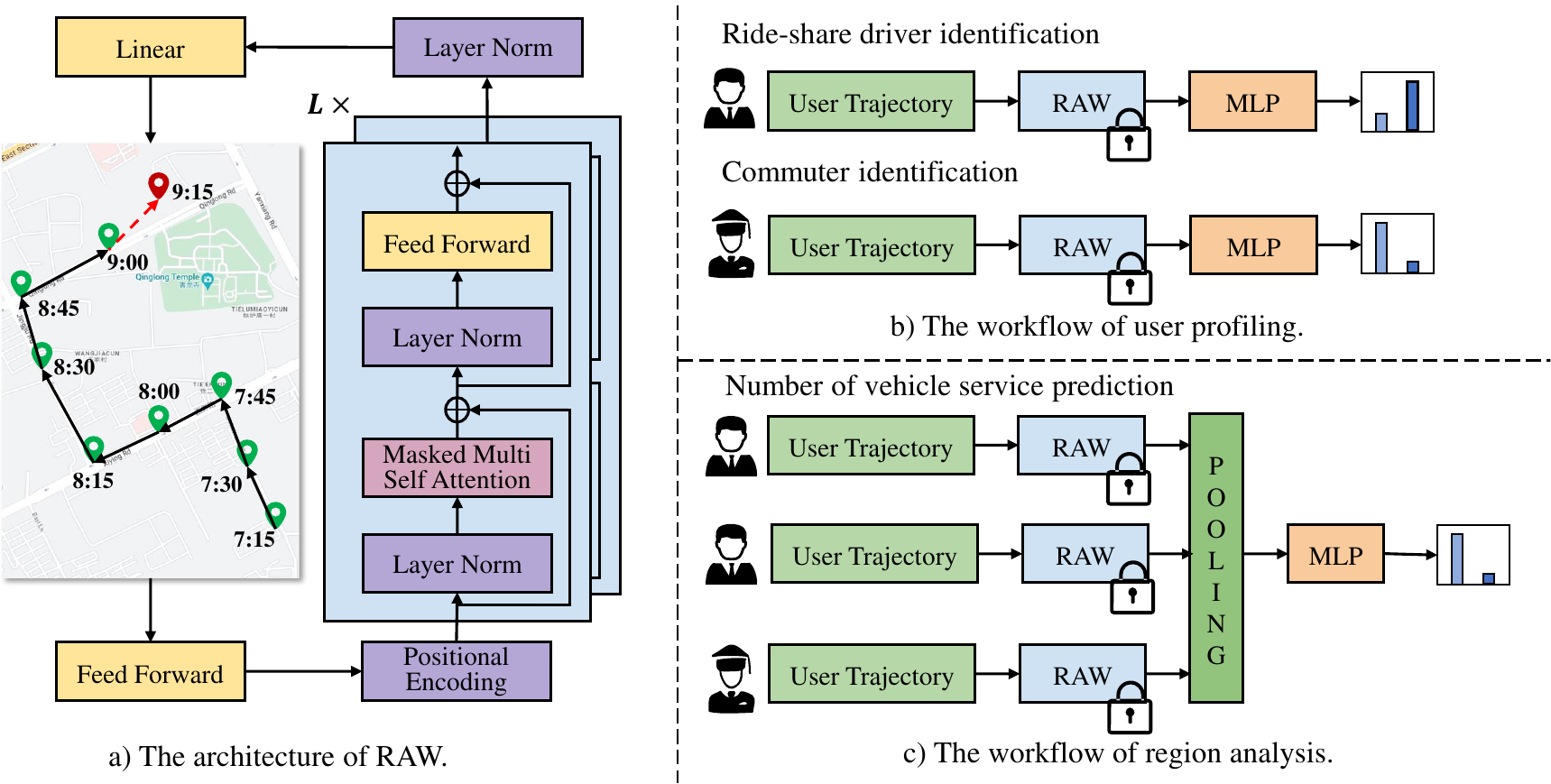}
  \caption{The framework of RAW. RAW first uses trajectories to train a GPT-like framework to obtain the embedding for each trajectory. When encountering a user profiling task, we use users' trajectory embedding as users' embedding. Then an extra MLP layer is trained to adapt the users' embedding to specific tasks. When encountering a region analysis task, we use aggregated users' embedding as regions' embedding. Then an extra MLP layer is trained to adapt the regions' embedding to specific tasks. }
  \label{fig2}
\end{figure*}

\section{Model}

Our model can be divided into two stages: the training phase and the fine-tuning phase.

In the training phase, our objective is to learn embeddings of trajectories through a self-supervised approach. Specifically, we begin by encoding each GPS coordinate point in the trajectory to obtain a GPS embedding. To allow the model to capture the sequential information in trajectories, we incorporate positional encoding with these encoded embeddings. Subsequently, these embeddings with positional encoding are fed into a network comprising L layers of transformer decoder layers. Each transformer decoder layer consists of sub-layers such as layer normalization\cite{layer_norm}, self-attention mechanisms, and feed-forward layers. Finally, the trajectory embedding we seek is obtained by inputting the GPT network's output embedding into a extra layer normalization layer. To train this network, we employ a task that predicts the next GPS coordinate point. This involves feeding the trajectory embedding into a linear layer to obtain the predicted value of the subsequent GPS coordinate point. It should be noticed that this framework only differs with GPT-3 in the extra feed forward layer at the input.

In the fine-tuning phase, we adapt and optimize the trajectory embedding according to different downstream tasks. For user profiling tasks, we directly input the user's trajectory embedding into a MLP to generate predictions for the user profile. In region analysis tasks, we initially extract region-specific portions from trajectory embeddings of users who have traversed that region, perform pooling operations to obtain region embeddings, and subsequently input these region embeddings into an MLP to generate predictions for the region analysis. We will then proceed to detail each module.

\subsection{Feed Forward Networks}
Before feeding the trajectory into the transformer decoder layer, we first need to encode GPS as a structured vector. Thus, we utilize a fully connected feed-forward network, which consists of two linear layers with a GELU activation function in between. Besides, a dropout layer will be added after the feed forward network\cite{dropout}.
\begin{equation}
    \text{FFN}(\mathbf X) = \text{GeLU}(\mathbf X\mathbf W_1 + \mathbf b_1)\mathbf W_2 + \mathbf b_2
\end{equation}
which $\mathbf W_1\in  R^{2\times 4d}$ and $\mathbf W_2\in R^{4d\times d}$. Besides, this structure will also be used in the transformer decoder layer to thoroughly extract information within the embedding. In that case, $\mathbf W_1\in R^{d\times 4d}$ and $\mathbf W_2\in R^{4d\times d}$, where $d$ represents the dimension of the embedding.

\subsection{Positional Encoding}

To enable our model to be sensitive to the sequence order, we need to incorporate positional encoding into the model. Here, we directly learn a positional encoding matrix, $\mathbf{W}_p$, to encode the $i^{th}$ item in the sequence into a d-dimensional vector. Overall, we first subject the input sequence T to the following transformation:
\begin{equation}
    \mathbf{H}^0 = \text{FFN}(T)+\mathbf{W}_p
\end{equation}
where $\mathbf{H}_0$ is the input of the transformer decoder layers.

\subsection{Multi Head Self-Attention}

In order to enable the model to perceive the information within the trajectory, we employ an attention mechanism to compute weighted sums over the sequence. The principle of the attention mechanism involves utilizing a vector to query a set of key vectors to obtain a weight. Then the weight is used to sum up a set of value vectors to obtain the output. Specifically, we first perform a dot product operation between the query vector and all key vectors, followed by scaling by $\sqrt{d}$, and then applying a softmax function to obtain weight coefficients for each value vector. Finally, the weighted coefficients are multiplied with the value vectors and summed, resulting in the output of the attention mechanism, which can be represented mathematically as:
\begin{equation}
\text{Attention}(\mathbf Q,\mathbf K,\mathbf V)=\text{softmax}(\frac{\mathbf Q\mathbf K^T}{\sqrt{d}})\mathbf V
\end{equation}
Where $\mathbf Q$ represents the query vector, $\mathbf K$ denotes the key vector matrix, $\mathbf V$ signifies the value vector matrix.

In order for the model to capture information from multiple perspectives, this paper also utilizes a multi-head attention mechanism, which can be defined as:
\begin{equation}
\begin{aligned}
    \text{MultiHead}(\mathbf Q,\mathbf K,\mathbf V) &= \text{Concat}(\text{head}_1, \text{head}_2, ..., \text{head}_h)\mathbf W^O \\
    where\ \text{head}_i &= \text{Attention}(\mathbf Q\mathbf W_i^Q,\mathbf K \mathbf W_i^K,\mathbf V\mathbf W_i^V)
\end{aligned}
\end{equation}
which $\mathbf W_i^*\in R^{d \times d_h}$, $d_h = d / h$ and $h$ is the number of the heads.

\subsection{Transformer Decoder Layers}
As shown in Figure \ref{fig2}, the transformer decoder layer is composed of the following four sub-layers in sequence: Layer norm, Masked Multi-Head Self-Attention, Layer Norm, Feed Forward. Among these, the multi-head self-attention is modified in a masked form to prevent the model from acquiring future information during prediction. The specific workflow of these four sub-layers is as follows: First, the input $\mathbf{H}^l$ undergoes layer normalization and masked multi-head self-attention, resulting in a new vector. To ensure gradient propagation and network depth, we employ a residual connection by adding the input $\mathbf{H}^l$ to the output of self-attention, yielding $\tilde{\mathbf{H}}^l$. Subsequently, $\tilde{\mathbf{H}}^l$ undergoes layer normalization and feed-forward layers, producing another new vector. Similarly, we apply a residual connection by adding $\tilde{\mathbf{H}}^l$ to the output of the feed-forward network, obtaining $\mathbf{H}^{l+1}$. After stacking the transformer decoder layer L times, an additional layer normalization is applied to achieve the desired trajectory embedding. This procedure can be defined as:
\begin{equation}
\begin{aligned}
    \mathbf{H}^{l*} &= \text{LayerNorm}(\mathbf{H}^l) \\
    \tilde{\mathbf{H}}^l &= \mathbf{H}^l + \text{MultiHead}(\mathbf{H}^{l*},\mathbf{H}^{l*},\mathbf{H}^{l*})
\end{aligned}
\end{equation}
\begin{equation}
    \mathbf{H}^{l+1} = \tilde{\mathbf{H}}^l + \text{FFN}(\text{LayerNorm}(\mathbf{H}^{l*}))
\end{equation}
\begin{equation}
    \mathbf{E} = \text{LayerNorm}(\mathbf{H}^L)
\end{equation}

\subsection{Output}

Taking inspiration from the GPT series, which employs the task of predicting the next word to train language models, we also adopted the task of predicting the next GPS coordinate point to train our trajectory model. To accomplish this task, we input the trajectory embedding $\mathbf e$ obtained from the transformer decoder layers into a linear layer, resulting in the predicted value of the next GPS coordinate point, represented mathematically as:
\begin{equation}    
\hat{x}_{i,|T_i|+1},\hat{y}_{i,|T_i|+1} = \mathbf{e}_i\mathbf W_o+\mathbf{b}_o
\end{equation}
Here, $W_o\in \mathbb{R}^{d\times 2}$, $\hat{x}_{i,|T_i|+1},\hat{y}_{i,|T_i|+1}$ represent the predicted coordinates of the $(|T_i|+1)$-th point of the $i$-th trajectory, and $\mathbf{b}_o$ denotes the bias term of the linear layer.

\subsection{User Profiling Fine-Tuning}
One of our fundamental assumption is that \textit{You are where you walk}. Thus, we directly utilize a user's trajectory embedding as the representation of a user. Specifically, for tasks such as identifying commuters, we input the user's trajectory into the trained RAW structure, obtaining the user's embedding $\mathbf e_i$. Subsequently, we train an additional MLP layer to determine whether the user is a commuter. Formally, for user $i$ and task $j$, we can obtain the prediction using the following formula:
\begin{equation}
    \begin{aligned}
        \mathbf e_i  &= \text{RAW}(T_i)\\
               p_{i,j} &= \text{MLP}_j(\mathbf e_i)
    \end{aligned}   
\end{equation}
where, $\text{MLP}_j$ represents the MLP layer to train for task $j$, and $p_{i,j}$ denotes the probability that user $i$ belongs to the target group for task $j$.

\subsection{Region Analysis Fine-Tuning}

One of our fundamental assumption is that \textit{Regions are who walk them}. Therefore, we directly employ the embedding of users who have traversed or been in proximity to that region as the embedding of the region itself. Specifically, for tasks such as functional zone identification in an region analysis, we begin by extracting trajectories of users who have passed through or have been closest to the area within the last K days. These trajectories are then fed into the trained RAW structure, generating a collection of user embeddings $\{\mathbf e_i\}$. Subsequently, the collection undergoes a sum pooling operation to derive the region embedding $\mathbf g_a$ . This method is employed because we believe that summation can reflect both the number of users passing through the region and retain the characteristic information of those users. Following this, we train an additional MLP layer to predict the functionality of the region. Formally, for region $a$ and functional zone $j$, we obtain the prediction using the following formula:
\begin{equation}
    \begin{aligned}
         \{\mathbf e_i\}  &= \text{RAW}(\{T_i\}) \\
           \mathbf g_a  &= \text{SUM}(\{\mathbf e_i\})\\
               p_{a,j} &= \text{MLP}_j(\mathbf g_a)
    \end{aligned}   
\end{equation}
where $\text{MLP}_j$ represents the MLP layer to train for identifying functional zone $j$, and $p_{a,j}$ signifies the probability that region $a$ belongs to functional zone $j$. 

\begin{figure}[t]
  \centering
  \includegraphics[width=0.9\linewidth]{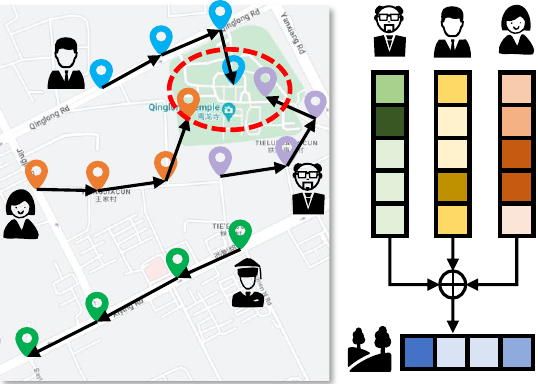}
  \caption{The illustration of region analysis fine-tuning. When generating initial embedding for the park(the green area in the picture), we first collect all users' embedding who have been or are close to this region in previous K days. Then, the embeddings are summed up to generate the initial embedding of the park. Further, this initial embedding will be used as the input of a task-specified MLP to make predictions. }
  \label{fig3}
\end{figure}

\section{Experiment}

\subsection{Experiment Setup}

\subsubsection{Datasets}
We trained RAW on the China Mobile Cellular Data, which consists of about 1 million cellular trajectories in Beijing. Each cellular trajectory contains a user’s cellular communication records in 10 days. For user profiling and region analysis, we collect a classification dataset and regression dataset for each of the task.

\textbf{Commuter} For user profiling tasks in classification, we select about 25,000 people who are commuters and 25,000 people who are not commuters by a predefined DTW\cite{DTW}-like algorithm. 

\textbf{Trip Count} For user profiling tasks in regression, we select about 25,000 people who take subway more than or equal to once in the given 10 days. Then we predict the number of taking subway in the 10 days using the person's trajectory.

\textbf{POI Most Type} For region analysis in classification, we chose 14 different POI types and calculate their corresponding number in different regions. This task aims at predicting which POI has the most number in a given region.

\textbf{Car Selling Service} For region analysis in regression, we calculate the number of car selling service in different regions. This task aims at predicting the number of the car selling service in a given region.

\subsubsection{Model Configurations}
We successfully trained RAW in four different settings on 8 NVIDIA A100 with PyTorch 2.0\cite{pytorch}, FSDP\cite{FSDP} and Adam Optimizer\cite{Adam}. The details can be found in Table \ref{tab1}. We only report the performance of RAW-tiny, while the performance of other settings will be released in a future version.

\begin{table}[h]
    \centering
    \begin{tabular}{ccccc}
        \toprule
        Model Name & $n_{params}$ & $n_{layers}$ & $d$ & $n_{heads}$\\
        \midrule
        RAW-tiny & 0.1B & 12 & 768 & 12\\
        RAW-small & 0.3B & 24 & 1024 & 16\\
        RAW-middle & 0.7B & 24 & 1536 & 16\\
        RAW-large & 1.3B & 24 & 2048 & 24\\
        \bottomrule
    \end{tabular}
    \caption{The model configuration of RAW.}
    \label{tab1}
\end{table}

\subsubsection{Evaluation Metrices}
We use three popular evaluation metrics to evaluate the downstream regression tasks: Mean Absolute Error (MAE), Root Mean Square Error (RMSE), and Pearson’s Correlation Coefficient (PCC). And we use another two popular evaluation metrics to evaluate the downstream classification tasks: Accuracy Score (ACC) and Hamming loss (Hamming).

\subsubsection{Baselines}
Since our basic assumption is that we do not have any prior features on users or regions, we only conducted two very simple baselines: Unique and Statics.

\textbf{Unique}: This baseline uses the most frequently occurring category in the training set as the prediction result for classification tasks, and the mean value of the training set for regression tasks. Therefore, this model predicts the same value, which is the best hypothesis made without any feature assumptions.

\textbf{Statics}: This is a baseline for region analysis that we statistically derive a 240-dimensional representation containing features such as traffic flow based on the existing data. We input this representation into a MLP layer for training and prediction. Therefore, this model simulates some related works that use population flow information for prediction.

\subsection{Model Performance}

\subsubsection{User Profiling Performance} We employed regression and classification tasks to assess the effectiveness of user profiling. For regression tasks, compared to using mean-based methods, we achieved reductions of 1 unit in MAE and  2 unit in RMSE. Additionally, we attained a PCC of 0.7, indicating a certain degree of correlation. In the case of classification tasks, our proposed method achieved over 85\% accuracy in identifying commuter users. These results demonstrate the strong capabilities of our representations in uncovering user attributes and patterns.

\begin{table}[H]
\begin{tabular}{c|ccc|cc}
\toprule
         & \multicolumn{3}{c|}{Trip Count} & \multicolumn{2}{c}{Commuter} \\ \midrule
Model    & MAE        & RMSE       & PCC       & ACC                & Haming                \\ \midrule
Unique  &     3.8318	&6.4493&	-&	0.5276&	0.4723\\
RAW-tiny &   2.6750 &	 3.9061&0.7120	 &	0.8374&	0.1625                   \\ \bottomrule

\end{tabular}
\caption{Performance of User Profiling}
\end{table}

\subsubsection{Region Analysis Performance} We also employed regression and classification tasks to evaluate the effectiveness of region analysis. In regression tasks, compared to mean-based methods, we achieved reductions of over 4 units in both MAE and RMSE. Even when using statistics-based features, achieving satisfactory performance in regression tasks was challenging, with PCC nearly zero. However, with our method, PCC could reach 0.6, indicating higher correlation. In classification tasks, our method also achieved the highest accuracy. Nevertheless, due to the complex nature of region attributes, directly determining the most frequent category posed significant randomness, resulting in relatively lower accuracy for both statistical and our proposed methods. These comparable results of our model highlight the robust capabilities of our representations in uncovering region attributes and patterns.

\begin{table}[H]
\begin{tabular}{c|ccc|cc}
\toprule
         & \multicolumn{3}{c|}{Car Selling Service} & \multicolumn{2}{c}{POI Most Type} \\ \midrule
Model    & MAE        & RMSE       & PCC       & ACC                & Haming                \\ \midrule

Unique  &    10.6992 & 18.5566 & 0	&	0.0694 & 0.9305\\
Statics  &     8.7090 &	19.6609 &	0.0214 & 0.2045 & 0.7954\\
RAW-tiny &  6.4660 & 13.6987 & 0.6764 & 0.2879 & 0.712 \\ 
\bottomrule
\end{tabular}
\caption{Performance of Region Analysis}
\end{table}

\subsubsection{Trajectory Prediction Performance} 
\begin{figure}[h]
	\begin{subfigure}{0.49\linewidth}
 		\centering
		\includegraphics[width=\linewidth]{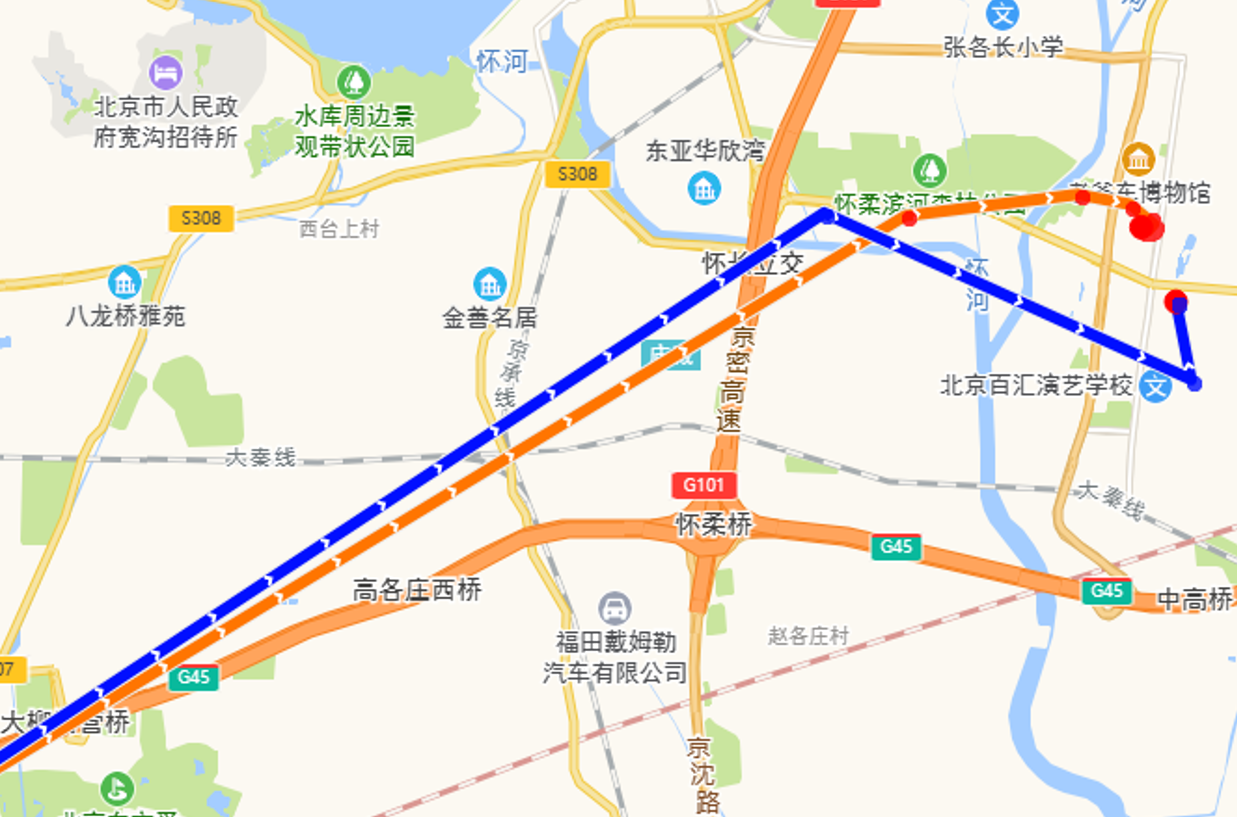}
            \caption{Trajectory Prediction Case 1}

	\end{subfigure}
	\begin{subfigure}{0.49\linewidth}
		\centering
		\includegraphics[width=\linewidth]{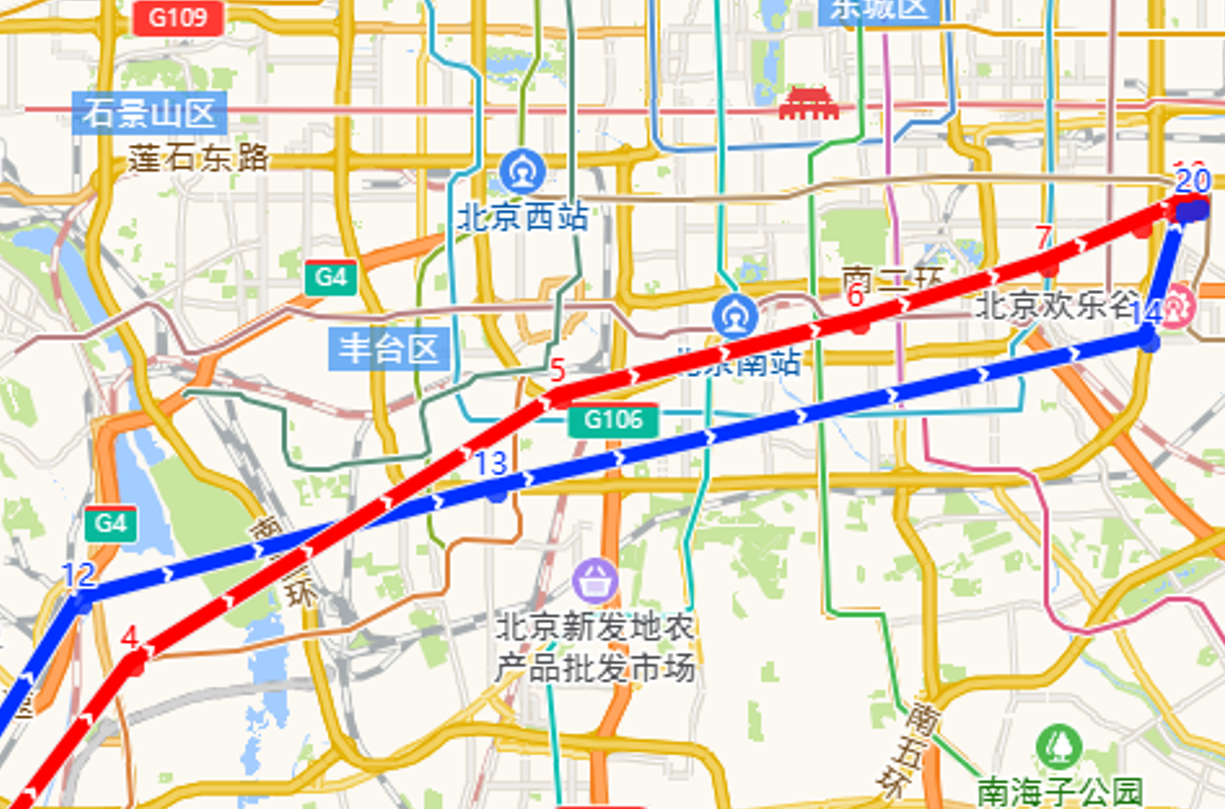}
		\caption{Trajectory Prediction Case 2}
	\end{subfigure}
     \caption{Trajectory Prediction Illustration. The blue one is the real trajectory while the red one is the predicted trajectory.}
	\label{fig4}
\end{figure}

As our training task involves predicting the next GPS coordinate point, our model can progressively predict future trajectories. To evaluate our ability, we designed an experiment where we used a trajectory with 950 coordinate points to generatively predict the subsequent 10 points. The experimental results in \ref{tab5} revealed that even when predicting positions in 2.5 hours, the individual longitude or latitude error did not exceed 1.5 kilometers. The average distance error for the GPS coordinate points also remained under 2.5 kilometers. These data substantiate our effective learning of user travel patterns. Additionally, through visual experiments such as the example shown in Figure \ref{fig4}, we observed that our model accurately predicts travel trajectories, speed, and destinations, further illustrating the meaningful embeddings obtained from the upstream pre-training task.

\begin{table}[h]
    \centering
    \begin{tabular}{c|cccc}
    \toprule
        Future & 08:00:00 & 11:00:00 & 14:00:00 & 17:00:00 \\ 
    \midrule
        0.25h & 333.2417 & 340.9027 & 360.9573 & 337.8745  \\
        0.50h & 465.6523 & 491.0523 & 504.9300 & 490.7477 \\ 
        0.75h & 561.2326 & 603.8546 & 635.5276 & 614.3375 \\ 
        1.00h & 685.9954 & 722.9634 & 770.3080 & 723.8778  \\ 
        1.25h & 782.7354 & 845.6154 & 908.3913 & 818.6221 \\
        1.50h & 867.0064 & 964.1990 & 1022.3561 & 871.2852 \\
        1.75h & 940.4992 & 1062.5358 & 1132.8988 & 928.1951 \\
        2.00h & 1052.3637 & 1155.7303 & 1270.8835 & 960.5879 \\ 
        2.25h & 1113.3307 & 1267.3995 & 1387.2078 & 987.2139 \\ 
        2.50h & 1180.4494 & 1354.8497 & 1478.7375 & 1015.9421 \\ 
    \bottomrule
    \end{tabular}
     \caption{Trajectory Prediction Result}
     \label{tab5}
\end{table}

\section{Conclusion and Future Work}

In this paper, we propose RAW, a large spatiotemporal model based on human mobility that can be scaled to 1B to realize user profiling and region analysis.
By modifying the GPT structure, we have successfully developed a model capable of generative pre-training for trajectories, thereby obtaining embeddings of trajectories.
To achieve user profiling, we directly employ a user's trajectory embedding as the user's embeddings and train an additional MLP layer to derive the predicted results for user profiling.
For region analysis, we utilize the aggregated embeddings of users who have traversed or been near a specific area as the area's embedding. We further train an additional MLP layer to obtain the predicted results for region analysis.
Extensive experiments have validated our predictive performance over baselines across multiple tasks in user profiling and region analysis.
Furthermore, in tasks involving trajectory generation, it is evident that our model has learned information regarding people's travel patterns, residential and working areas.

In the future, we have two valuable research directions.
The first is trajectory generation based on prompts. This involves providing a user's historical trajectories and designing some prompting trajectories to simulate travel or traffic events.
The second involves mobility simulation based on the large model. Due to the sensitivity of user trajectory data, which cannot be publicly shared, generating a simulation dataset using large models could circumvent privacy issues and support researchers in their studies.

\bibliographystyle{ACM-Reference-Format}
\bibliography{sample-base}


\begin{thebibliography}{24}


\ifx \showCODEN    \undefined \def \showCODEN     #1{\unskip}     \fi
\ifx \showDOI      \undefined \def \showDOI       #1{#1}\fi
\ifx \showISBNx    \undefined \def \showISBNx     #1{\unskip}     \fi
\ifx \showISBNxiii \undefined \def \showISBNxiii  #1{\unskip}     \fi
\ifx \showISSN     \undefined \def \showISSN      #1{\unskip}     \fi
\ifx \showLCCN     \undefined \def \showLCCN      #1{\unskip}     \fi
\ifx \shownote     \undefined \def \shownote      #1{#1}          \fi
\ifx \showarticletitle \undefined \def \showarticletitle #1{#1}   \fi
\ifx \showURL      \undefined \def \showURL       {\relax}        \fi
\providecommand\bibfield[2]{#2}
\providecommand\bibinfo[2]{#2}
\providecommand\natexlab[1]{#1}
\providecommand\showeprint[2][]{arXiv:#2}

\bibitem[Ba et~al\mbox{.}(2016)]%
        {layer_norm}
\bibfield{author}{\bibinfo{person}{Lei~Jimmy Ba}, \bibinfo{person}{Jamie~Ryan Kiros}, {and} \bibinfo{person}{Geoffrey~E. Hinton}.} \bibinfo{year}{2016}\natexlab{}.
\newblock \showarticletitle{Layer Normalization}.
\newblock \bibinfo{journal}{\emph{CoRR}}  \bibinfo{volume}{abs/1607.06450} (\bibinfo{year}{2016}).
\newblock
\showeprint[arXiv]{1607.06450}
\urldef\tempurl%
\url{http://arxiv.org/abs/1607.06450}
\showURL{%
\tempurl}


\bibitem[Batran et~al\mbox{.}(2018)]%
        {human_mobility_data}
\bibfield{author}{\bibinfo{person}{Mohamed Batran}, \bibinfo{person}{Mariano~Gregorio Mejia}, \bibinfo{person}{Yoshihide Sekimoto}, {and} \bibinfo{person}{Ryosuke Shibasaki}.} \bibinfo{year}{2018}\natexlab{}.
\newblock \showarticletitle{Inference of Human Spatiotemporal Mobility in Greater Maputo by Mobile Phone Big Data Mining}. In \bibinfo{booktitle}{\emph{Proceedings of the Tenth International Workshop on Agents in Traffic and Transportation {(ATT} 2018) co-located with with the Federated Artificial Intelligence Meeting, including ECAI/IJCAI, {AAMAS} and {ICML} 2018 conferences {(FAIM} 2018), Stockholm, Sweden, July 14, 2018}} \emph{(\bibinfo{series}{{CEUR} Workshop Proceedings}, Vol.~\bibinfo{volume}{2129})}, \bibfield{editor}{\bibinfo{person}{Ana L{\'{u}}cia~C. Bazzan}, \bibinfo{person}{Luca Crociani}, \bibinfo{person}{Ivana Dusparic}, {and} \bibinfo{person}{Sascha Ossowski}} (Eds.). \bibinfo{publisher}{CEUR-WS.org}, \bibinfo{pages}{1--8}.
\newblock
\urldef\tempurl%
\url{https://ceur-ws.org/Vol-2129/paper3.pdf}
\showURL{%
\tempurl}


\bibitem[Berndt and Clifford(1994)]%
        {DTW}
\bibfield{author}{\bibinfo{person}{Donald~J. Berndt} {and} \bibinfo{person}{James Clifford}.} \bibinfo{year}{1994}\natexlab{}.
\newblock \showarticletitle{Using Dynamic Time Warping to Find Patterns in Time Series}. In \bibinfo{booktitle}{\emph{Knowledge Discovery in Databases: Papers from the 1994 {AAAI} Workshop, Seattle, Washington, USA, July 1994. Technical Report {WS-94-03}}}, \bibfield{editor}{\bibinfo{person}{Usama~M. Fayyad} {and} \bibinfo{person}{Ramasamy Uthurusamy}} (Eds.). \bibinfo{publisher}{{AAAI} Press}, \bibinfo{pages}{359--370}.
\newblock


\bibitem[Brown et~al\mbox{.}(2020)]%
        {GPT-3}
\bibfield{author}{\bibinfo{person}{Tom~B. Brown}, \bibinfo{person}{Benjamin Mann}, \bibinfo{person}{Nick Ryder}, \bibinfo{person}{Melanie Subbiah}, \bibinfo{person}{Jared Kaplan}, \bibinfo{person}{Prafulla Dhariwal}, \bibinfo{person}{Arvind Neelakantan}, \bibinfo{person}{Pranav Shyam}, \bibinfo{person}{Girish Sastry}, \bibinfo{person}{Amanda Askell}, \bibinfo{person}{Sandhini Agarwal}, \bibinfo{person}{Ariel Herbert{-}Voss}, \bibinfo{person}{Gretchen Krueger}, \bibinfo{person}{Tom Henighan}, \bibinfo{person}{Rewon Child}, \bibinfo{person}{Aditya Ramesh}, \bibinfo{person}{Daniel~M. Ziegler}, \bibinfo{person}{Jeffrey Wu}, \bibinfo{person}{Clemens Winter}, \bibinfo{person}{Christopher Hesse}, \bibinfo{person}{Mark Chen}, \bibinfo{person}{Eric Sigler}, \bibinfo{person}{Mateusz Litwin}, \bibinfo{person}{Scott Gray}, \bibinfo{person}{Benjamin Chess}, \bibinfo{person}{Jack Clark}, \bibinfo{person}{Christopher Berner}, \bibinfo{person}{Sam McCandlish}, \bibinfo{person}{Alec Radford}, \bibinfo{person}{Ilya Sutskever},
  {and} \bibinfo{person}{Dario Amodei}.} \bibinfo{year}{2020}\natexlab{}.
\newblock \showarticletitle{Language Models are Few-Shot Learners}.
\newblock \bibinfo{journal}{\emph{CoRR}}  \bibinfo{volume}{abs/2005.14165} (\bibinfo{year}{2020}).
\newblock
\showeprint[arXiv]{2005.14165}
\urldef\tempurl%
\url{https://arxiv.org/abs/2005.14165}
\showURL{%
\tempurl}


\bibitem[Driess et~al\mbox{.}(2023)]%
        {multimodal1}
\bibfield{author}{\bibinfo{person}{Danny Driess}, \bibinfo{person}{Fei Xia}, \bibinfo{person}{Mehdi S.~M. Sajjadi}, \bibinfo{person}{Corey Lynch}, \bibinfo{person}{Aakanksha Chowdhery}, \bibinfo{person}{Brian Ichter}, \bibinfo{person}{Ayzaan Wahid}, \bibinfo{person}{Jonathan Tompson}, \bibinfo{person}{Quan Vuong}, \bibinfo{person}{Tianhe Yu}, \bibinfo{person}{Wenlong Huang}, \bibinfo{person}{Yevgen Chebotar}, \bibinfo{person}{Pierre Sermanet}, \bibinfo{person}{Daniel Duckworth}, \bibinfo{person}{Sergey Levine}, \bibinfo{person}{Vincent Vanhoucke}, \bibinfo{person}{Karol Hausman}, \bibinfo{person}{Marc Toussaint}, \bibinfo{person}{Klaus Greff}, \bibinfo{person}{Andy Zeng}, \bibinfo{person}{Igor Mordatch}, {and} \bibinfo{person}{Pete Florence}.} \bibinfo{year}{2023}\natexlab{}.
\newblock \showarticletitle{PaLM-E: An Embodied Multimodal Language Model}. In \bibinfo{booktitle}{\emph{International Conference on Machine Learning, {ICML} 2023, 23-29 July 2023, Honolulu, Hawaii, {USA}}} \emph{(\bibinfo{series}{Proceedings of Machine Learning Research}, Vol.~\bibinfo{volume}{202})}, \bibfield{editor}{\bibinfo{person}{Andreas Krause}, \bibinfo{person}{Emma Brunskill}, \bibinfo{person}{Kyunghyun Cho}, \bibinfo{person}{Barbara Engelhardt}, \bibinfo{person}{Sivan Sabato}, {and} \bibinfo{person}{Jonathan Scarlett}} (Eds.). \bibinfo{publisher}{{PMLR}}, \bibinfo{pages}{8469--8488}.
\newblock
\urldef\tempurl%
\url{https://proceedings.mlr.press/v202/driess23a.html}
\showURL{%
\tempurl}


\bibitem[Gong et~al\mbox{.}(2023)]%
        {multimodal2}
\bibfield{author}{\bibinfo{person}{Tao Gong}, \bibinfo{person}{Chengqi Lyu}, \bibinfo{person}{Shilong Zhang}, \bibinfo{person}{Yudong Wang}, \bibinfo{person}{Miao Zheng}, \bibinfo{person}{Qian Zhao}, \bibinfo{person}{Kuikun Liu}, \bibinfo{person}{Wenwei Zhang}, \bibinfo{person}{Ping Luo}, {and} \bibinfo{person}{Kai Chen}.} \bibinfo{year}{2023}\natexlab{}.
\newblock \showarticletitle{MultiModal-GPT: {A} Vision and Language Model for Dialogue with Humans}.
\newblock \bibinfo{journal}{\emph{CoRR}}  \bibinfo{volume}{abs/2305.04790} (\bibinfo{year}{2023}).
\newblock
\urldef\tempurl%
\url{https://doi.org/10.48550/ARXIV.2305.04790}
\showDOI{\tempurl}
\showeprint[arXiv]{2305.04790}


\bibitem[Kingma and Ba(2015)]%
        {Adam}
\bibfield{author}{\bibinfo{person}{Diederik~P. Kingma} {and} \bibinfo{person}{Jimmy Ba}.} \bibinfo{year}{2015}\natexlab{}.
\newblock \showarticletitle{Adam: {A} Method for Stochastic Optimization}. In \bibinfo{booktitle}{\emph{3rd International Conference on Learning Representations, {ICLR} 2015, San Diego, CA, USA, May 7-9, 2015, Conference Track Proceedings}}, \bibfield{editor}{\bibinfo{person}{Yoshua Bengio} {and} \bibinfo{person}{Yann LeCun}} (Eds.).
\newblock
\urldef\tempurl%
\url{http://arxiv.org/abs/1412.6980}
\showURL{%
\tempurl}


\bibitem[Li et~al\mbox{.}(2023)]%
        {region_embed_1}
\bibfield{author}{\bibinfo{person}{Yi Li}, \bibinfo{person}{Weiming Huang}, \bibinfo{person}{Gao Cong}, \bibinfo{person}{Hao Wang}, {and} \bibinfo{person}{Zheng Wang}.} \bibinfo{year}{2023}\natexlab{}.
\newblock \showarticletitle{Urban Region Representation Learning with OpenStreetMap Building Footprints}. In \bibinfo{booktitle}{\emph{Proceedings of the 29th {ACM} {SIGKDD} Conference on Knowledge Discovery and Data Mining, {KDD} 2023, Long Beach, CA, USA, August 6-10, 2023}}, \bibfield{editor}{\bibinfo{person}{Ambuj~K. Singh}, \bibinfo{person}{Yizhou Sun}, \bibinfo{person}{Leman Akoglu}, \bibinfo{person}{Dimitrios Gunopulos}, \bibinfo{person}{Xifeng Yan}, \bibinfo{person}{Ravi Kumar}, \bibinfo{person}{Fatma Ozcan}, {and} \bibinfo{person}{Jieping Ye}} (Eds.). \bibinfo{publisher}{{ACM}}, \bibinfo{pages}{1363--1373}.
\newblock
\urldef\tempurl%
\url{https://doi.org/10.1145/3580305.3599538}
\showDOI{\tempurl}


\bibitem[Liang et~al\mbox{.}(2018)]%
        {user_profiling_3}
\bibfield{author}{\bibinfo{person}{Shangsong Liang}, \bibinfo{person}{Xiangliang Zhang}, \bibinfo{person}{Zhaochun Ren}, {and} \bibinfo{person}{Evangelos Kanoulas}.} \bibinfo{year}{2018}\natexlab{}.
\newblock \showarticletitle{Dynamic Embeddings for User Profiling in Twitter}. In \bibinfo{booktitle}{\emph{Proceedings of the 24th {ACM} {SIGKDD} International Conference on Knowledge Discovery {\&} Data Mining, {KDD} 2018, London, UK, August 19-23, 2018}}, \bibfield{editor}{\bibinfo{person}{Yike Guo} {and} \bibinfo{person}{Faisal Farooq}} (Eds.). \bibinfo{publisher}{{ACM}}, \bibinfo{pages}{1764--1773}.
\newblock
\urldef\tempurl%
\url{https://doi.org/10.1145/3219819.3220043}
\showDOI{\tempurl}


\bibitem[Paszke et~al\mbox{.}(2019)]%
        {pytorch}
\bibfield{author}{\bibinfo{person}{Adam Paszke}, \bibinfo{person}{Sam Gross}, \bibinfo{person}{Francisco Massa}, \bibinfo{person}{Adam Lerer}, \bibinfo{person}{James Bradbury}, \bibinfo{person}{Gregory Chanan}, \bibinfo{person}{Trevor Killeen}, \bibinfo{person}{Zeming Lin}, \bibinfo{person}{Natalia Gimelshein}, \bibinfo{person}{Luca Antiga}, \bibinfo{person}{Alban Desmaison}, \bibinfo{person}{Andreas K{\"{o}}pf}, \bibinfo{person}{Edward~Z. Yang}, \bibinfo{person}{Zach DeVito}, \bibinfo{person}{Martin Raison}, \bibinfo{person}{Alykhan Tejani}, \bibinfo{person}{Sasank Chilamkurthy}, \bibinfo{person}{Benoit Steiner}, \bibinfo{person}{Lu Fang}, \bibinfo{person}{Junjie Bai}, {and} \bibinfo{person}{Soumith Chintala}.} \bibinfo{year}{2019}\natexlab{}.
\newblock \showarticletitle{PyTorch: An Imperative Style, High-Performance Deep Learning Library}.
\newblock \bibinfo{journal}{\emph{CoRR}}  \bibinfo{volume}{abs/1912.01703} (\bibinfo{year}{2019}).
\newblock
\showeprint[arXiv]{1912.01703}
\urldef\tempurl%
\url{http://arxiv.org/abs/1912.01703}
\showURL{%
\tempurl}


\bibitem[Radford et~al\mbox{.}(2021)]%
        {CLIP}
\bibfield{author}{\bibinfo{person}{Alec Radford}, \bibinfo{person}{Jong~Wook Kim}, \bibinfo{person}{Chris Hallacy}, \bibinfo{person}{Aditya Ramesh}, \bibinfo{person}{Gabriel Goh}, \bibinfo{person}{Sandhini Agarwal}, \bibinfo{person}{Girish Sastry}, \bibinfo{person}{Amanda Askell}, \bibinfo{person}{Pamela Mishkin}, \bibinfo{person}{Jack Clark}, \bibinfo{person}{Gretchen Krueger}, {and} \bibinfo{person}{Ilya Sutskever}.} \bibinfo{year}{2021}\natexlab{}.
\newblock \showarticletitle{Learning Transferable Visual Models From Natural Language Supervision}. In \bibinfo{booktitle}{\emph{Proceedings of the 38th International Conference on Machine Learning, {ICML} 2021, 18-24 July 2021, Virtual Event}} \emph{(\bibinfo{series}{Proceedings of Machine Learning Research}, Vol.~\bibinfo{volume}{139})}, \bibfield{editor}{\bibinfo{person}{Marina Meila} {and} \bibinfo{person}{Tong Zhang}} (Eds.). \bibinfo{publisher}{{PMLR}}, \bibinfo{pages}{8748--8763}.
\newblock
\urldef\tempurl%
\url{http://proceedings.mlr.press/v139/radford21a.html}
\showURL{%
\tempurl}


\bibitem[Radford et~al\mbox{.}(2018)]%
        {GPT-1}
\bibfield{author}{\bibinfo{person}{Alec Radford}, \bibinfo{person}{Karthik Narasimhan}, \bibinfo{person}{Tim Salimans}, \bibinfo{person}{Ilya Sutskever}, {et~al\mbox{.}}} \bibinfo{year}{2018}\natexlab{}.
\newblock \showarticletitle{Improving language understanding by generative pre-training}.
\newblock  (\bibinfo{year}{2018}).
\newblock


\bibitem[Radford et~al\mbox{.}(2019)]%
        {GPT-2}
\bibfield{author}{\bibinfo{person}{Alec Radford}, \bibinfo{person}{Jeffrey Wu}, \bibinfo{person}{Rewon Child}, \bibinfo{person}{David Luan}, \bibinfo{person}{Dario Amodei}, \bibinfo{person}{Ilya Sutskever}, {et~al\mbox{.}}} \bibinfo{year}{2019}\natexlab{}.
\newblock \showarticletitle{Language models are unsupervised multitask learners}.
\newblock \bibinfo{journal}{\emph{OpenAI blog}} \bibinfo{volume}{1}, \bibinfo{number}{8} (\bibinfo{year}{2019}), \bibinfo{pages}{9}.
\newblock


\bibitem[Ramesh et~al\mbox{.}(2022)]%
        {dalle2}
\bibfield{author}{\bibinfo{person}{Aditya Ramesh}, \bibinfo{person}{Prafulla Dhariwal}, \bibinfo{person}{Alex Nichol}, \bibinfo{person}{Casey Chu}, {and} \bibinfo{person}{Mark Chen}.} \bibinfo{year}{2022}\natexlab{}.
\newblock \showarticletitle{Hierarchical Text-Conditional Image Generation with {CLIP} Latents}.
\newblock \bibinfo{journal}{\emph{CoRR}}  \bibinfo{volume}{abs/2204.06125} (\bibinfo{year}{2022}).
\newblock
\urldef\tempurl%
\url{https://doi.org/10.48550/ARXIV.2204.06125}
\showDOI{\tempurl}
\showeprint[arXiv]{2204.06125}


\bibitem[Rong et~al\mbox{.}(2020)]%
        {protein}
\bibfield{author}{\bibinfo{person}{Yu Rong}, \bibinfo{person}{Yatao Bian}, \bibinfo{person}{Tingyang Xu}, \bibinfo{person}{Weiyang Xie}, \bibinfo{person}{Ying Wei}, \bibinfo{person}{Wenbing Huang}, {and} \bibinfo{person}{Junzhou Huang}.} \bibinfo{year}{2020}\natexlab{}.
\newblock \showarticletitle{Self-Supervised Graph Transformer on Large-Scale Molecular Data}. In \bibinfo{booktitle}{\emph{Advances in Neural Information Processing Systems 33: Annual Conference on Neural Information Processing Systems 2020, NeurIPS 2020, December 6-12, 2020, virtual}}, \bibfield{editor}{\bibinfo{person}{Hugo Larochelle}, \bibinfo{person}{Marc'Aurelio Ranzato}, \bibinfo{person}{Raia Hadsell}, \bibinfo{person}{Maria{-}Florina Balcan}, {and} \bibinfo{person}{Hsuan{-}Tien Lin}} (Eds.).
\newblock
\urldef\tempurl%
\url{https://proceedings.neurips.cc/paper/2020/hash/94aef38441efa3380a3bed3faf1f9d5d-Abstract.html}
\showURL{%
\tempurl}


\bibitem[Srivastava et~al\mbox{.}(2014)]%
        {dropout}
\bibfield{author}{\bibinfo{person}{Nitish Srivastava}, \bibinfo{person}{Geoffrey~E. Hinton}, \bibinfo{person}{Alex Krizhevsky}, \bibinfo{person}{Ilya Sutskever}, {and} \bibinfo{person}{Ruslan Salakhutdinov}.} \bibinfo{year}{2014}\natexlab{}.
\newblock \showarticletitle{Dropout: a simple way to prevent neural networks from overfitting}.
\newblock \bibinfo{journal}{\emph{J. Mach. Learn. Res.}} \bibinfo{volume}{15}, \bibinfo{number}{1} (\bibinfo{year}{2014}), \bibinfo{pages}{1929--1958}.
\newblock
\urldef\tempurl%
\url{https://doi.org/10.5555/2627435.2670313}
\showDOI{\tempurl}


\bibitem[Vaswani et~al\mbox{.}(2017)]%
        {transformer}
\bibfield{author}{\bibinfo{person}{Ashish Vaswani}, \bibinfo{person}{Noam Shazeer}, \bibinfo{person}{Niki Parmar}, \bibinfo{person}{Jakob Uszkoreit}, \bibinfo{person}{Llion Jones}, \bibinfo{person}{Aidan~N. Gomez}, \bibinfo{person}{Lukasz Kaiser}, {and} \bibinfo{person}{Illia Polosukhin}.} \bibinfo{year}{2017}\natexlab{}.
\newblock \showarticletitle{Attention is All you Need}. In \bibinfo{booktitle}{\emph{Advances in Neural Information Processing Systems 30: Annual Conference on Neural Information Processing Systems 2017, December 4-9, 2017, Long Beach, CA, {USA}}}, \bibfield{editor}{\bibinfo{person}{Isabelle Guyon}, \bibinfo{person}{Ulrike von Luxburg}, \bibinfo{person}{Samy Bengio}, \bibinfo{person}{Hanna~M. Wallach}, \bibinfo{person}{Rob Fergus}, \bibinfo{person}{S.~V.~N. Vishwanathan}, {and} \bibinfo{person}{Roman Garnett}} (Eds.). \bibinfo{pages}{5998--6008}.
\newblock
\urldef\tempurl%
\url{https://proceedings.neurips.cc/paper/2017/hash/3f5ee243547dee91fbd053c1c4a845aa-Abstract.html}
\showURL{%
\tempurl}


\bibitem[Wang and Li(2017)]%
        {region_rep_1}
\bibfield{author}{\bibinfo{person}{Hongjian Wang} {and} \bibinfo{person}{Zhenhui Li}.} \bibinfo{year}{2017}\natexlab{}.
\newblock \showarticletitle{Region Representation Learning via Mobility Flow}. In \bibinfo{booktitle}{\emph{Proceedings of the 2017 {ACM} on Conference on Information and Knowledge Management, {CIKM} 2017, Singapore, November 06 - 10, 2017}}, \bibfield{editor}{\bibinfo{person}{Ee{-}Peng Lim}, \bibinfo{person}{Marianne Winslett}, \bibinfo{person}{Mark Sanderson}, \bibinfo{person}{Ada~Wai{-}Chee Fu}, \bibinfo{person}{Jimeng Sun}, \bibinfo{person}{J.~Shane Culpepper}, \bibinfo{person}{Eric Lo}, \bibinfo{person}{Joyce~C. Ho}, \bibinfo{person}{Debora Donato}, \bibinfo{person}{Rakesh Agrawal}, \bibinfo{person}{Yu~Zheng}, \bibinfo{person}{Carlos Castillo}, \bibinfo{person}{Aixin Sun}, \bibinfo{person}{Vincent~S. Tseng}, {and} \bibinfo{person}{Chenliang Li}} (Eds.). \bibinfo{publisher}{{ACM}}, \bibinfo{pages}{237--246}.
\newblock
\urldef\tempurl%
\url{https://doi.org/10.1145/3132847.3133006}
\showDOI{\tempurl}


\bibitem[Wang et~al\mbox{.}(2019)]%
        {user_profiling_2}
\bibfield{author}{\bibinfo{person}{Pengyang Wang}, \bibinfo{person}{Yanjie Fu}, \bibinfo{person}{Hui Xiong}, {and} \bibinfo{person}{Xiaolin Li}.} \bibinfo{year}{2019}\natexlab{}.
\newblock \showarticletitle{Adversarial Substructured Representation Learning for Mobile User Profiling}. In \bibinfo{booktitle}{\emph{Proceedings of the 25th {ACM} {SIGKDD} International Conference on Knowledge Discovery {\&} Data Mining, {KDD} 2019, Anchorage, AK, USA, August 4-8, 2019}}, \bibfield{editor}{\bibinfo{person}{Ankur Teredesai}, \bibinfo{person}{Vipin Kumar}, \bibinfo{person}{Ying Li}, \bibinfo{person}{R{\'{o}}mer Rosales}, \bibinfo{person}{Evimaria Terzi}, {and} \bibinfo{person}{George Karypis}} (Eds.). \bibinfo{publisher}{{ACM}}, \bibinfo{pages}{130--138}.
\newblock
\urldef\tempurl%
\url{https://doi.org/10.1145/3292500.3330869}
\showDOI{\tempurl}


\bibitem[Wang et~al\mbox{.}(2020)]%
        {user_profiling_1}
\bibfield{author}{\bibinfo{person}{Pengyang Wang}, \bibinfo{person}{Kunpeng Liu}, \bibinfo{person}{Lu Jiang}, \bibinfo{person}{Xiaolin Li}, {and} \bibinfo{person}{Yanjie Fu}.} \bibinfo{year}{2020}\natexlab{}.
\newblock \showarticletitle{Incremental Mobile User Profiling: Reinforcement Learning with Spatial Knowledge Graph for Modeling Event Streams}. In \bibinfo{booktitle}{\emph{{KDD} '20: The 26th {ACM} {SIGKDD} Conference on Knowledge Discovery and Data Mining, Virtual Event, CA, USA, August 23-27, 2020}}, \bibfield{editor}{\bibinfo{person}{Rajesh Gupta}, \bibinfo{person}{Yan Liu}, \bibinfo{person}{Jiliang Tang}, {and} \bibinfo{person}{B.~Aditya Prakash}} (Eds.). \bibinfo{publisher}{{ACM}}, \bibinfo{pages}{853--861}.
\newblock
\urldef\tempurl%
\url{https://doi.org/10.1145/3394486.3403128}
\showDOI{\tempurl}


\bibitem[Wu et~al\mbox{.}(2022)]%
        {region_rep_2}
\bibfield{author}{\bibinfo{person}{Shangbin Wu}, \bibinfo{person}{Xu Yan}, \bibinfo{person}{Xiaoliang Fan}, \bibinfo{person}{Shirui Pan}, \bibinfo{person}{Shichao Zhu}, \bibinfo{person}{Chuanpan Zheng}, \bibinfo{person}{Ming Cheng}, {and} \bibinfo{person}{Cheng Wang}.} \bibinfo{year}{2022}\natexlab{}.
\newblock \showarticletitle{Multi-Graph Fusion Networks for Urban Region Embedding}. In \bibinfo{booktitle}{\emph{Proceedings of the Thirty-First International Joint Conference on Artificial Intelligence, {IJCAI} 2022, Vienna, Austria, 23-29 July 2022}}, \bibfield{editor}{\bibinfo{person}{Luc~De Raedt}} (Ed.). \bibinfo{publisher}{ijcai.org}, \bibinfo{pages}{2312--2318}.
\newblock
\urldef\tempurl%
\url{https://doi.org/10.24963/IJCAI.2022/321}
\showDOI{\tempurl}


\bibitem[Zhang et~al\mbox{.}(2020)]%
        {region_rep_3}
\bibfield{author}{\bibinfo{person}{Mingyang Zhang}, \bibinfo{person}{Tong Li}, \bibinfo{person}{Yong Li}, {and} \bibinfo{person}{Pan Hui}.} \bibinfo{year}{2020}\natexlab{}.
\newblock \showarticletitle{Multi-View Joint Graph Representation Learning for Urban Region Embedding}. In \bibinfo{booktitle}{\emph{Proceedings of the Twenty-Ninth International Joint Conference on Artificial Intelligence, {IJCAI} 2020}}, \bibfield{editor}{\bibinfo{person}{Christian Bessiere}} (Ed.). \bibinfo{publisher}{ijcai.org}, \bibinfo{pages}{4431--4437}.
\newblock
\urldef\tempurl%
\url{https://doi.org/10.24963/IJCAI.2020/611}
\showDOI{\tempurl}


\bibitem[Zhang et~al\mbox{.}(2023)]%
        {region_rep_4}
\bibfield{author}{\bibinfo{person}{Yu Zhang}, \bibinfo{person}{Yonghui Xu}, \bibinfo{person}{Lizhen Cui}, {and} \bibinfo{person}{Zhongmin Yan}.} \bibinfo{year}{2023}\natexlab{}.
\newblock \showarticletitle{Multi-View Graph Contrastive Learning for Urban Region Representation}. In \bibinfo{booktitle}{\emph{International Joint Conference on Neural Networks, {IJCNN} 2023, Gold Coast, Australia, June 18-23, 2023}}. \bibinfo{publisher}{{IEEE}}, \bibinfo{pages}{1--8}.
\newblock
\urldef\tempurl%
\url{https://doi.org/10.1109/IJCNN54540.2023.10191432}
\showDOI{\tempurl}


\bibitem[Zhao et~al\mbox{.}(2023)]%
        {FSDP}
\bibfield{author}{\bibinfo{person}{Yanli Zhao}, \bibinfo{person}{Andrew Gu}, \bibinfo{person}{Rohan Varma}, \bibinfo{person}{Liang Luo}, \bibinfo{person}{Chien{-}Chin Huang}, \bibinfo{person}{Min Xu}, \bibinfo{person}{Less Wright}, \bibinfo{person}{Hamid Shojanazeri}, \bibinfo{person}{Myle Ott}, \bibinfo{person}{Sam Shleifer}, \bibinfo{person}{Alban Desmaison}, \bibinfo{person}{Can Balioglu}, \bibinfo{person}{Bernard Nguyen}, \bibinfo{person}{Geeta Chauhan}, \bibinfo{person}{Yuchen Hao}, {and} \bibinfo{person}{Shen Li}.} \bibinfo{year}{2023}\natexlab{}.
\newblock \showarticletitle{PyTorch {FSDP:} Experiences on Scaling Fully Sharded Data Parallel}.
\newblock \bibinfo{journal}{\emph{CoRR}}  \bibinfo{volume}{abs/2304.11277} (\bibinfo{year}{2023}).
\newblock
\urldef\tempurl%
\url{https://doi.org/10.48550/ARXIV.2304.11277}
\showDOI{\tempurl}
\showeprint[arXiv]{2304.11277}


\end{thebibliography}

\end{document}